\documentclass[letterpaper, 10 pt, conference]{ieeeconf}  

\IEEEoverridecommandlockouts                              

\usepackage[T1]{fontenc}
\usepackage{blindtext}
\usepackage{graphicx}
\usepackage{xcolor,colortbl}

\usepackage{amsmath}
\usepackage{amssymb}

\usepackage{graphicx}
\usepackage{placeins}

\usepackage{subfig}
\usepackage{gensymb}
\usepackage{textcomp}
\usepackage{amsmath} 
\usepackage{diagbox}

\usepackage{amsmath}

\usepackage{mathtools}

\usepackage{lipsum}

\usepackage{flushend}

\usepackage[colorinlistoftodos]{todonotes}

\usepackage{tikz}
\usepackage{textcomp}
\usepackage{hyperref}
\usepackage{lipsum}

\newcommand\copyrighttext{%
  \footnotesize \textcopyright 2020 IEEE. Personal use of this material is permitted.
  Permission from IEEE must be obtained for all other uses, in any current or future
  media, including reprinting/republishing this material for advertising or promotional
  purposes, creating new collective works, for resale or redistribution to servers or
  lists, or reuse of any copyrighted component of this work in other works.
  DOI: \href{to be added}{to be added}}
\newcommand\copyrightnotice{%
\begin{tikzpicture}[remember picture,overlay]
\node[anchor=south,yshift=10pt] at (current page.south) {\fbox{\parbox{\dimexpr\textwidth-\fboxsep-\fboxrule\relax}{\copyrighttext}}};
\end{tikzpicture}%
}

\title{\LARGE \bf
Comparison of camera-based and 3D LiDAR-based \\ loop closures across weather conditions
}

\author{Kamil \.Zywanowski*$^{1}$, Adam Banaszczyk*$^{1}$, Micha\l{} R. Nowicki$^{1}$
\thanks{* Equal contribution}
\thanks{$^{1}$ The authors are with the Institute of Robotics and Machine Intelligence,
Faculty of Control, Robotics and Electrical Engineering,
        Poznan University of Technology, Poznan, Poland
        {\tt\small michal.nowicki@put.poznan.pl}}%
}

\begin{document}

\maketitle
\copyrightnotice

\thispagestyle{empty}
\pagestyle{empty}

\begin{abstract}
Loop closure based on camera images provides excellent results on benchmarking datasets, but might struggle in real-world adverse weather conditions like direct sun, rain, fog, or just darkness at night.
In automotive applications, the sensory setups include 3D LiDARs that provide information complementary to cameras. 
The presented article focuses on the evaluation of camera-based, LiDAR-based, and joint camera-LiDAR-based loop closures applying a similar processing pipeline consisting of a neural network under varying weather conditions using the newly available USyd dataset.
The experiments performed on the same trajectories in diverse weather conditions over 50 weeks prove that a 16-line 3D LiDAR can be used to supplement image-based loop closure to increase loop closure performance.
This proves that there is a need for more research into loop closures performed with multi-sensory setups.
\end{abstract}

\section{Introduction}


The key element of effective long-time robot localization is the ability to reduce the accumulated drift when a robot revisits an already known location~\cite{tutorialPerception}.
The so-called loop closure can be performed based on a variety of sensors with the GPS and the camera being the prime examples. 
The GPS signal is sometimes unavailable and therefore, the appearance-based loop closure is used to determine place similarity solely on its visual characteristic and without prior geometric assumptions.

The systems to detect loop closures using images from RGB cameras are already used in real-world scenarios~\cite{fabmap,dbow,seqslam}.
The performance of these methods depends on the image quality that degenerates at night, in adverse weather conditions, or when the sun shines directly into the lens blinding the camera.
Fortunately, most of the autonomous cars are equipped with other sensors like 3D LiDARs to provide necessary robustness in these conditions.
The scans from 3D LiDARs provide information about the geometry of the surroundings of the robot, complementing RGB images from cameras to form a more complete view of the environment.

\begin{figure}[htbp!]
    \centering
    \includegraphics[width=\columnwidth]{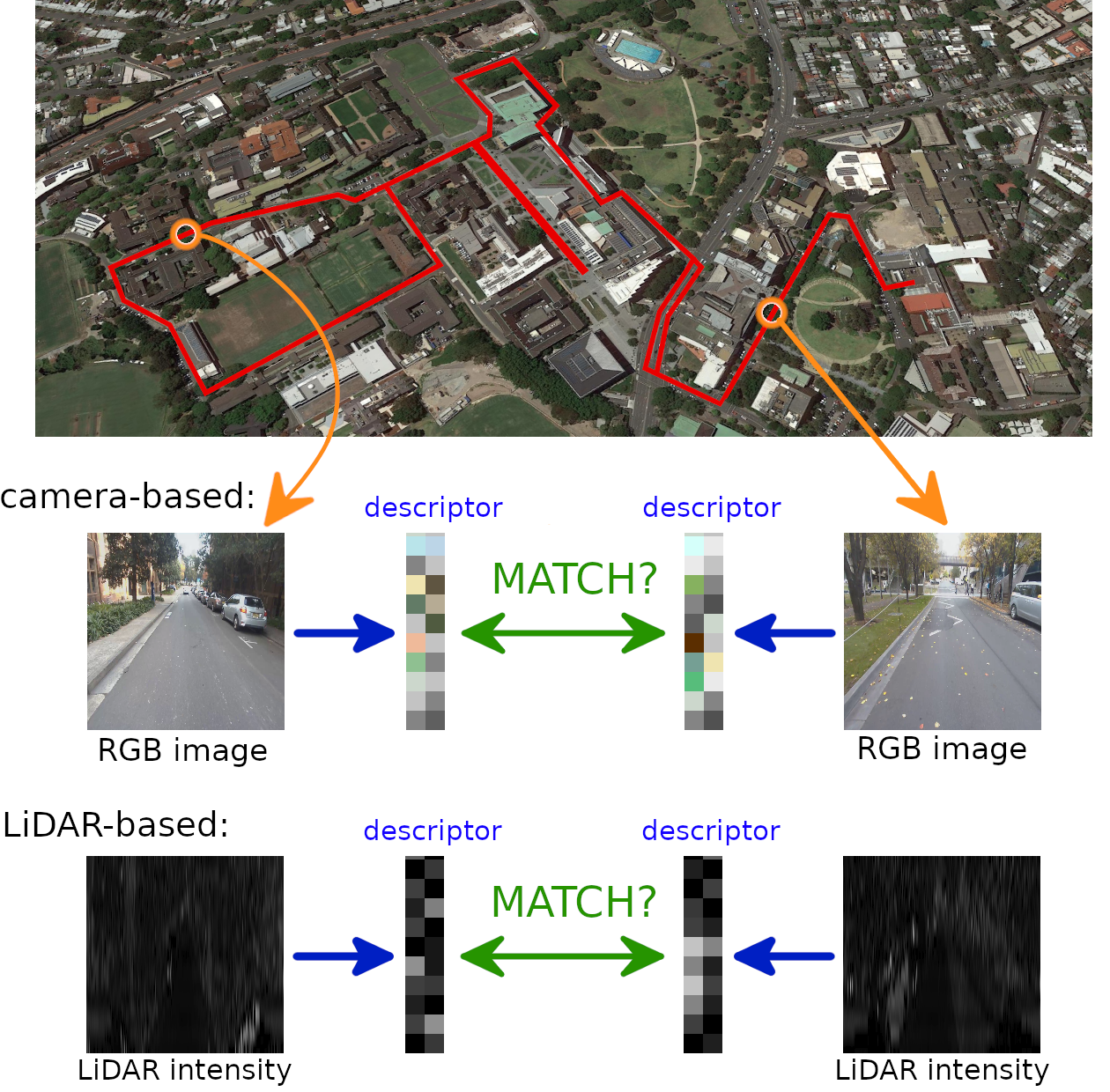}
    \caption{We compare loop closure performed on descriptors trained on RGB images (camera-based), LiDAR intensity values (LiDAR-based) and joint input of RGB images and LiDAR intensity values (camera-LiDAR-based) in varying weather conditions on multiple runs on the same trajectory using USyD dataset.
    }
    \label{fig:catchy_image}
\end{figure}

In contrary to camera-based loop closure, the problem of loop closure using LiDAR data is still actively researched with most current efforts focusing on proper point cloud representation for deep learning~\cite{seqlpd,pointnetvlad,lpdnet}.
To the best knowledge of the authors, the joint RGB-LiDAR loop closure is still an unexplored research direction. 
The goal of the presented article is to determine the real-world conditions in which the camera-based, LiDAR-based, and camera-LiDAR-based loop closures provide satisfactory or poor results (Fig.~\ref{fig:catchy_image}) providing first camera-LiDAR-based loop closure pipeline.
In order to achieve this goal, we utilize the University of Sydney Campus Dataset~\cite{usyd} that provides camera images with 16-line LiDAR that was gathered across varying weather conditions.
Our processing pipeline, based on~\cite{singleview,multiview}, is used with minor modifications for all considered versions to focus on performance under changing weather conditions while reducing the influence that could stem from different processing pipelines.

The contribution of our work can be summarized as:
\begin{itemize}
    \item the first experimental verification of the LiDAR-based loop closure in changing weather conditions.
    \item the first experimental comparison between camera-based and LiDAR-based loop closures using similar processing pipelines on the same sequences.
    \item the first multi-sensory camera-LiDAR-based loop closure system with extensive experimental verification.
\end{itemize}

\section{Related work}

The appearance-based loop closure using RGB images from cameras is a well-researched topic with several established solutions that can be divided into two groups of approaches.
The first is based on constructing a global descriptor from local features, like FABMAP~\cite{fabmap} or DBoW~\cite{dbow}.
These methods use the global descriptor with the bag of visual words (BoVW) approach to determine the similarity of locations usually based on data from a single location.
The second group is based on utilizing simpler and faster to compute global descriptors, but relying on sequences of these descriptors to achieve the desired efficiency, like SeqSLAM~\cite{seqslam} or FastABLE~\cite{wpc}.
The advent of CNN resulted in approaches with learnable features that are more robust to weather conditions or lightning changes, i.e. as in the work of Naseer \textit{et al.}~\cite{burgardRGB}.

Due to the increasing popularity of LiDARs in the automotive industry, the 3D LiDARs are getting cheaper while at the same time, the typical number of scan lines increases, resulting in a denser representation of the environment and new application possibilities, i.e. to use them for loop closure. 
Before the advent of deep learning, the feature-based approaches to LiDAR-based solutions were popular with either specific local interest points~\cite{depthLFeat1, depthLFeat2} or global frame description~\cite{depthGFeat1}.
In the deep learning era, we see further improvements in local~\cite{burgardDepth} and global~\cite{ishot} descriptors.
Notably, in~\cite{ishot}, authors propose a descriptor that joints depth measurements with returning signal intensity to propose a globally invariant place descriptor.
Nevertheless, the most articles on the point cloud-based loop closure focus on point cloud representation for deep learning that is used to train the descriptor, i.e. as in~\cite{seqlpd}. 
The PointNet representation with NetVLAD as in~\cite{pointnetvlad} or 
graph-based neighborhood aggregation as in~\cite{lpdnet} can be used, but there seems to still be room for improvement with better point cloud representations.
On the other hand, the SegMatch~\cite{segmatch} avoids the problem of proper point cloud representation by matching hand-crafted descriptors of the segmented parts of a point cloud. 

In our comparison, similarly to SegMatch, we wanted to avoid the problem of proper point cloud representation for deep learning to keep our pipeline similar for RGB and LiDAR data. 
Therefore, following remarks highlighting the importance of LiDAR intensity and its invariance to lighting conditions~\cite{ishot}, we focus only on the LiDAR intensity information ignoring depth measurements and utilizing 2D image representation for intensity measurements.
Our approach is based on RGB image descriptor learning with triplet loss, as in~\cite{multiview}, that is applied in the same way for both RGB and LiDAR intensity input.
With such an approach, our processing pipeline is similar to~\cite{locnet}, where CNN on range image with depth measurements from LiDAR is trained with the contrastive loss to achieve robust place descriptor.

\section{Processing pipeline}

\begin{figure*}[h!]
 \centering
  \includegraphics[width=0.65\textwidth]{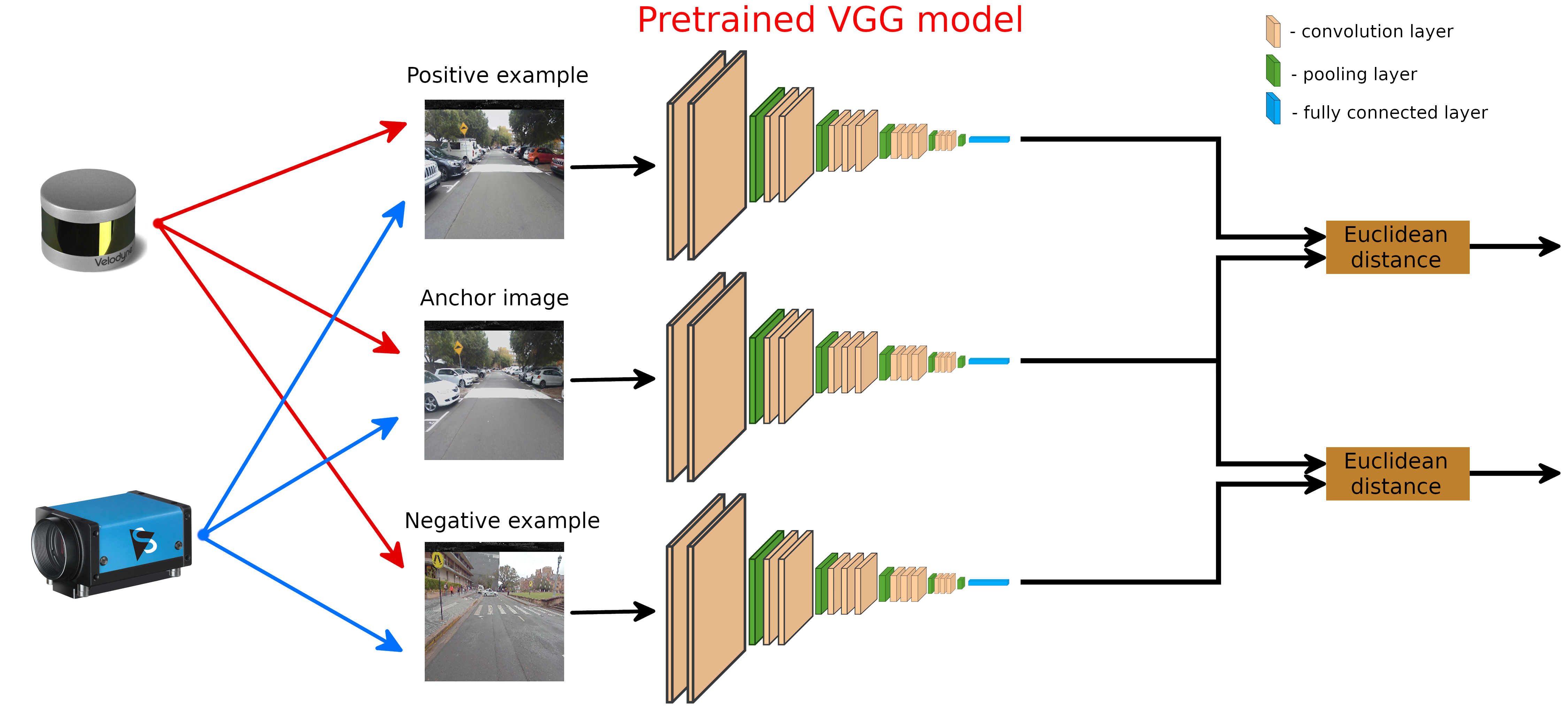}
    \caption{The descriptor for each location is the last layer of the deep neural network trained with triplet loss. The overall architecture of the networks is the same for camera-based, LiDAR-based, and joint camera-LiDAR-based loop closures with each solution trained on its own database}
    \label{fig:triplet_changed}
\end{figure*}

The network used in our comparison was proposed by F\'acil \textit{et al.}~\cite{singleview,multiview} and is presented in Fig.~\ref{fig:triplet_changed}.
The network consists of three identical processing pipelines that take three $224\times224$ pixel, 3-channel images as an input.
The training input is comprised of an anchor image (reference image), an image that is a positive match, and an image that is a negative match to the anchor image.
The network is trained simultaneously with positive and negative pairs, which makes the learning process more stable and efficient.
As a result of training, the descriptors obtained from the same place are getting more similar according to the chosen Euclidean norm, while the distance of the descriptors obtained from different places increases. 
The initial part of the network is a pre-trained part of the VGG-16 model extended by a fully-connected layer without activation function directly after the max-pooling layer.
For training, we use the Wohlhart-Lepetit loss (also called triplet loss) proposed in \cite{tripleloss}:

\begin{equation} \label{loss_equ}
    E = max \Bigg\{ 0.1 - \frac{d_n}{margin + d_p} \Bigg\}
\end{equation}

Where \textit{E} is the loss error, $d_n$ is the distance between the neutral and negative input, $d_p$ is the distance between the neutral and positive input and \textit{margin} is an additional parameter which helps to limit the difference between those two distances. 
In all the experiments the margin parameter was set to 1. The value of the loss is limited between 0 and 1, with a function returning 0 whenever the distance to the positive pair is smaller than the distance to the negative pair plus the margin.
The network at its final layer generates a concise descriptor of the place that is used to detect loop closures. 

In the case of the RGB images, we only resize the available images to fit the assumed input size of the image. 
In the case of the LiDAR, we only utilize the intensity channel and thus we decided to represent it as intensity images, also resized to fit the required input size.
In both cases our input is an image and thus we have a similar network pipeline for RGB and point cloud data to infer the influence of the weather conditions.

\section{Evaluation methodology}

The loop closure algorithms are usually evaluated on Nordlandsbanen~\cite{nordland}, Oxford RobotCar dataset~\cite{robotcar}, or KITTI dataset~\cite{kitti}. 
Neither of these datasets fits the requirements to perform a reliable comparison between camera-based and LiDAR-based loop closures using a location description from a single image or a single LiDAR scan.
We wanted to use a single instance of measurement to create a global descriptor that later can be extended with known DBoW or NetVLAD multi-place frameworks.
The Nordlandsbanen lacks the LiDAR, the Oxford RobotCar dataset contains only 4-layer LiDARs, and KITTI does not have enough varying conditions for a reliable comparison.
Even though the depth data from the RobotCar dataset could be made denser by combining multiple scans based on odometry, we wanted to achieve an independent global descriptor for each location without the necessity to stop to gather dense LiDAR scans, i.e. as performed in~\cite{denseLaser}.

\subsection{University of Sydney Campus Dataset}

In our comparison, we utilize the University of Sydney dataset (USyd)~\cite{usyd} that contains recordings of data collected by multiple sensors: cameras, 3D LiDAR (Velodyne VLP-16), u-blox GPS, IMU, and others while driving almost the same route once a week for more than one year. Currently, it consists of over 50 recordings covering different illumination and weather conditions as well as infrastructural, environmental, and traffic variations making it a perfect experimental setup to compare the camera-based and LiDAR-based loop closures.

In our comparison, we are interested in timestamped measurements from front-facing camera images, 3D LiDAR scans, and corresponding GPS measurements. 
The GPS data is converted into the local, metric coordinate system with UTM (Universal Transverse Mercator) conversion.
Since the sensors recorded information with different timestamps and different frequencies, the location from the GPS is linearly interpolated in metric coordinates to provide a location for each considered RGB image and LiDAR scan.
The role of the GPS information is to provide ground truth locations of the processed images and LiDAR scans and to make it possible to match measurements representing the same place from multiple recorded runs. 
The LiDAR scans used by our processing pipeline are not motion-compensated resulting in distortions while moving with greater speed similarly to distortions observed for moving rolling-shutter cameras.

\begin{figure}[h!]
 \centering
  \includegraphics[width=\columnwidth]{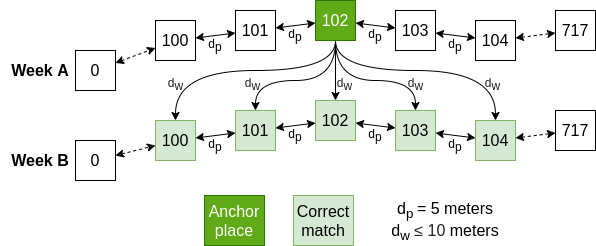}
    \caption{Locations chosen in the data are separated by $d_p = 5$ meters. We consider that two places from week A and week B are matching only if the distance between them does not exceed $d_w = 10$ meters}
    \label{fig:gps}
\end{figure}

We took the first recording and divided it into discrete places with each location separated by at least $d_p = 5$ meters from any other, as presented in Fig.~\ref{fig:gps}. 
We omitted the places with no GPS data available, as we would not be able use them in training, nor in verification.
In total, we obtained 718 distinct locations in the experimental environment. 
From these locations, we used 446 places for training and 163 places for testing.
To provide separation of training and testing locations, we rejected 109 places.

The route, on which the data was collected, in some parts, covers the same streets but in a different direction.
Despite the VLP-16 capability of recording the surrounding scene in its full range of 360\textdegree, we assumed that the same physical location with reverse heading is treated as another place to be able to perform a direct comparison with RGB-based solution.

\subsection{Training}

We used transfer learning to train our networks with the weights (coefficients) of the VGG-16 pre-trained on the Places database~\cite{places}. 
The same pre-trained model was used for RGB and LiDAR intensity images. 
The idea of this procedure is to use a trained model, which to some extent is capable of recognizing locations 
and adapt it for the new task. 

The proposed network is trained with a triplet loss that requires defining positive and negative examples. 
We consider two RGB images (or two intensity images from LiDAR) to be a positive example if the distance between their locations is within the threshold of $d_w = 10$ meters. The threshold was chosen based on the knowledge of the limited dynamic accuracy of the typical GPS. 
The pairs of data from the outside of that threshold are considered negative examples.
For each location determined in the USyd dataset, we found all of the images and LiDAR scans that fit within the assumed threshold and then generated positive pairs based on these matches.
Each positive pair has an associated negative pair that in the original database was chosen randomly as long as the distance to the anchor was greater than $t_n = 50$ meters.
Based on the USyD sequences, we obtained approximately 4.3 million triplets that were used for training.
The random choice of a negative example with $t_n = 50$ meters leads to a plateau in training as the number of non-active triplets greatly exceeds the number of active cases during later stages of training.
To overcome this issue, we prepared a separate database with hard negative examples, chosen with the distance to the anchor equal to $25$ meters.
The new database was used to train the network (fine-tune) once the training plateau on the original database was observed and proved to significantly increase our recognition accuracy by approximately $7\%$ on the testing set.

\section{Experimental results}

In the presented experiments, we assumed that the loop closure is trained on images but operates in an unknown environment and is expected to determine place similarity based on a single previous observation. 
With this assumption, the data from one week of the USyD dataset is treated as a reference and the data from another week is considered as testing to measure loop closure recognition accuracy. 
The original dataset consists of 52 weeks but only 38 of these contained correct 3D LiDAR, RGB images, and GPS data needed for reliable comparison.
Based on dataset author's annotations, we group the obtained results based on the weather conditions observed for each week into 6 categories: sunny (\textbf{S}), cloudy (\textbf{C}), sunny/cloudy (\textbf{S/C}), after rain (\textbf{AR}), sunset (\textbf{SS}), and very cloudy (\textbf{VC}). 
This clustering makes it possible to verify if and how the weather conditions influence the performance of the loop closure.

\begin{table}[htbp!]
\caption{Number of testing locations based on reference (Ref.) and testing (Test) week when divided into categories based on weather conditions: sunny (\textbf{S}), cloudy (\textbf{C}), sunny/cloudy (\textbf{S/C}), after rain (\textbf{AR}), sunset (\textbf{SS}), very cloudy (\textbf{VC})}
\label{tab:testsize}
\centering
\begin{tabular}{c|cccccc}
\diagbox{Test}{Ref.}    & \textbf{S} & \textbf{C} & \textbf{S/C} & \textbf{AR} & \textbf{SS} & \textbf{VC} \\ \hline
\textbf{S}   & 43738 & 16782 & 7771 & 9193 & 5063 & 2617 \\
\textbf{C}   & 16782 & 5468  & 2852 & 3275 & 1851 & 953  \\
\textbf{S/C} & 7771  & 2852  & 864  & 1539 & 854  & 442  \\
\textbf{AR}  & 9193  & 3275  & 1539 & 1318 & 1007 & 519  \\
\textbf{SS}  & 5063  & 1851  & 854  & 1007 & 288  & 287  \\
\textbf{VC}  & 2617  & 953   & 442  & 519  & 287  & 0   
\end{tabular}
\end{table}

Naturally, our dataset is not well-balanced, which can be observed by analyzing the number of testing locations in Tab.~\ref{tab:testsize}, and reflects the frequency of the conditions in the real-world.
As it reflects the real-world conditions, we do not perform any special actions to balance the distribution in our dataset.

\subsection{Camera-based loop closure}

The accuracy of the camera-based loop closure was verified on all of the available test locations.
We compared the descriptor of the testing location to all of the descriptors of the locations available in the reference. 
If the most similar location based on the similarity of the single-place descriptor was within $\pm 10$ meters of the location measured from the GPS, the found location was assumed to be correct.
In all other cases, the testing location was marked as incorrect.
The threshold of $\pm 10$ meters was chosen experimentally as a sufficient accuracy of the appearance-based solution that should converge to real metric localization if a geometric approach, like ICP~\cite{icp}, would be used.

\begin{table}[htbp!]
\centering
\caption{The recognition accuracy in percentages based on the reference (columns) and testing (rows) weather conditions for camera-based loop closure. Notice the lowered performance compared to average when testing in sunny conditions}
\label{tab:rgb}
\begin{tabular}{c|cccccc|c}
\textbf{Camera}  & \textbf{S} & \textbf{C} & \textbf{S/C} & \textbf{AR} & \textbf{SS} & \textbf{VC} & \textbf{Mean} \\ \hline
 \textbf{S}    & 80.32 & 83.20 & 85.39 & 84.61 & 83.33 & 83.03 & \cellcolor{red!25}82.08          \\
\textbf{C}    & 83.26 & 85.55 & 86.68 & 87.27 & 87.74 & 86.99 & 84.78          \\
\textbf{S/C}  & 86.00 & 87.24 & 87.73 & 89.99 & 88.06 & 88.46 & 86.98          \\
\textbf{AR}   & 83.67 & 85.74 & 89.02 & 90.14 & 89.87 & 86.71 & 85.53          \\
\textbf{SS}   & 79.50 & 84.39 & 83.37 & 89.77 & 90.63 & 83.28 & \cellcolor{red!25}82.39          \\
\textbf{VC}   & 83.15 & 84.78 & 88.24 & 88.25 & 86.41 &   -   & 84.68          \\ \hline
\textbf{Mean} & 81.82 & 84.37 & 86.15 & 86.47 & 85.66 & 84.72 & \textbf{83.49}
\end{tabular}
\end{table}

Based on all tests, the recognition accuracy of the RGB image loop closure was measured to be equal to $83.49\%$.
The exact performance depending on varying weather conditions is presented in Tab.~\ref{tab:rgb}.
The poorest performance, marked by red background color, was observed for sunny and sunset conditions.

\begin{figure}[htbp!]
    \centering
    \includegraphics[width=0.7\columnwidth]{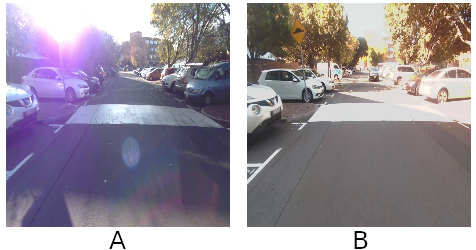}
    \caption{The visual comparison of images of the same location taken during sunset (A) and in sunny conditions (B) that are challenging for camera-based loop closure due to direct camera sunlight}
    \label{fig:rgbSun}
\end{figure}

In these cases, direct sunlight is a factor that can negatively influence the image acquisition process leading to overexposure that drastically changes the apparent perception of the location, i.e. as presented in Fig.~\ref{fig:rgbSun}.
There is no easy way to improve the quality of images from the chosen camera in such cases. In practice, the easiest way to deal with such situations is to use another camera that could be faced backward or rely on another type of sensor, like LiDAR.

\subsection{LiDAR-based loop closure}

Similarly to the camera-based loop closure, we also trained and analyzed the version operating on LiDAR intensities represented as an image. 
The overall successful recognition accuracy was measured to be equal to $81.11\%$, which is lower than the recognition accuracy reported for the camera-based solution.
We believe that it has to be expected as the Velodyne VLP-16 LiDAR used in the USyD dataset has only 16 independent horizontal lines that have to be significantly upscaled to match the expected input size of the network. 
In the case of the camera, the original image contains more independent information that has to be downsampled to fit the input of the network.
The more in-detail results across different weather conditions are presented in Tab.~\ref{tab:lidar}.

\begin{table}[htbp!]
\caption{The recognition accuracy in percentages based on the reference (columns) and testing (rows) weather conditions for LiDAR-based loop closure. Notice the overall similar performance apart from after rain (\textbf{AR}) conditions}
\label{tab:lidar}
\begin{tabular}{c|cccccc|c}
\textbf{LiDAR} & \textbf{S} & \textbf{C} & \textbf{S/C} & \textbf{AR} & \textbf{SS} & \textbf{VC} & \textbf{Mean}  \\ \hline
\textbf{S}    & 81.21 & 81.27 & 82.05 & 81.05 & 80.56 & 81.47 & \cellcolor{green!25}81.25          \\
\textbf{C}    & 81.22 & 78.58 & 80.29 & 78.69 & 79.15 & 79.64 & 80.24          \\
\textbf{S/C}  & 82.87 & 82.71 & 82.99 & 82.00 & 81.50 & 84.16 & 82.71          \\
\textbf{AR}   & 80.39 & 77.80 & 79.66 & 83.31 & 84.81 & 78.03 & \cellcolor{red!25}80.24          \\
\textbf{SS}   & 81.02 & 79.90 & 77.87 & 86.30 & 89.58 & 82.58 & \cellcolor{green!25}81.39          \\
\textbf{VC}   & 82.12 & 83.32 & 82.35 & 80.35 & 80.49 & -     & 82.09          \\\hline
\textbf{Mean} & 81.29 & 80.55 & 81.26 & 81.15 & 81.10 & 81.05 & \textbf{81.11}       \\
\end{tabular}
\end{table}

In most cases, the LiDAR-based solution performs similarly showing some robustness to weather conditions with a drop when the data acquisition was performed after raining (red background), which is expected as additional raindrops increase the number of missing measurements in the LiDAR data.
On the other hand, the LiDAR-based loop closure is robust to changes in lighting conditions working in sunny and sunset conditions (green background).

\begin{figure}[htbp!]
    \centering
    \includegraphics[width=\columnwidth]{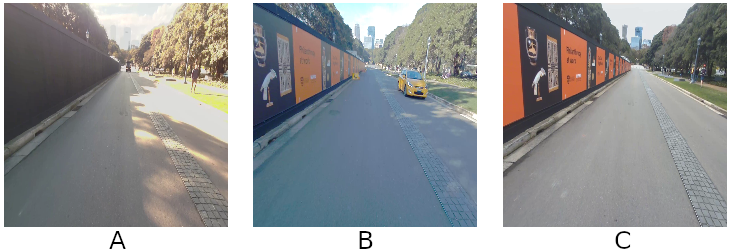}
    \caption{Example location visible at different times: with LiDAR-based solution working more reliably in case of more structure and intense sun (A), similar performance of camera- and LiDAR-based version in typical conditions (B) and camera-based version outperforming LiDAR in ideal lightning conditions (C)}
    \label{fig:rgblidar}
\end{figure}

Taking a closer look reveals that the performance of the LiDAR-based solution is more reliant on the geometry of the scene rather than the visual appearance.
Such a case is visible in Fig.~\ref{fig:rgblidar} when a lack of a poster on the wall misguides the camera-based loop closure but the LiDAR-based version is more robust.
Nevertheless, the LiDAR-based loop closure performs overall worse than the camera-based solution.

\subsection{Camera-LiDAR-based loop closure}

\begin{figure}[htbp!]
    \centering
    \includegraphics[width=\columnwidth]{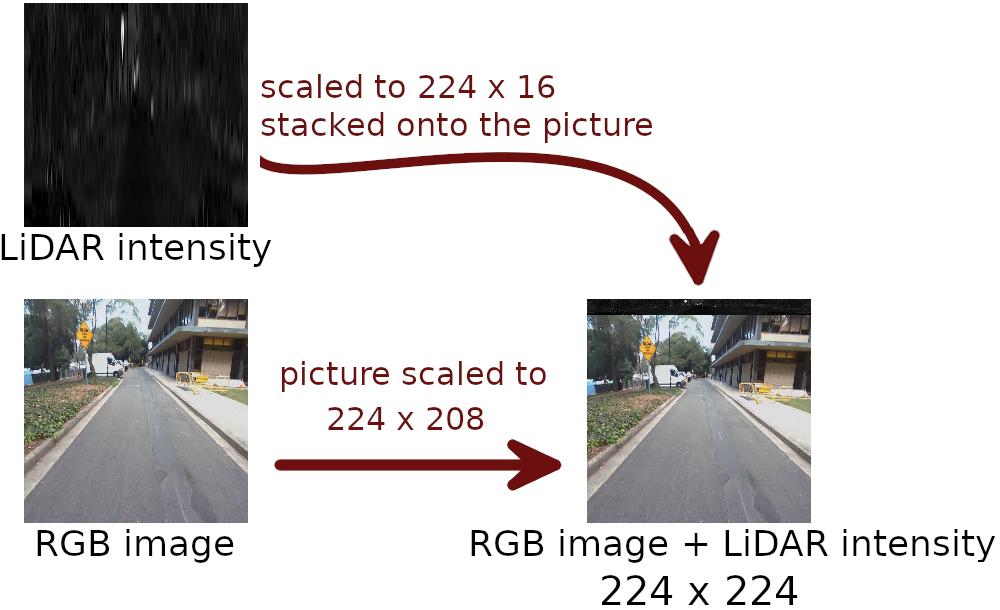}
    \caption{The visual representation of the camera-LiDAR-based image that is formed by joining resized LiDAR intensities with resized RGB image to form the artificial image. The artificial image has the same size as the inputs in the camera-based and LiDAR-based solutions}
    \label{fig:fusion}
\end{figure}

The results obtained from the camera-based loop closure could be improved when information from a sensor providing good performance in sunny and sunset situations could supplement the original data. 
Therefore, we verified the camera-LiDAR-based loop closure that was formed by joining a LiDAR intensity image with a camera image to form an artificial image.
The artificial image creation process is presented in Fig.~\ref{fig:fusion}.
In this artificial image, the first 16 rows contain the resized LiDAR intensity information, while the remaining 208 rows contain the resized RGB image.
Similarly to previous versions, we prepared a new training database, trained and then verified the performance of the network.

The camera-LiDAR-based loop closure achieved a correct recognition of $86.91\%$ on the testing sequence exceeding the results obtained from each of the individual sensors.
The exact performance in the analyzed weather conditions was measured and is presented in Tab.~\ref{tab:rgblidar}.

\begin{table}[htbp!]
\caption{The recognition accuracy in percentages based on the reference (columns) and testing (rows) weather conditions for camera-LiDAR-based loop closure. Notice the overall increase in the performance when compared to camera-based and LiDAR-based solutions}
\label{tab:rgblidar}
\centering
\begin{tabular}{c|cccccc|c}
\textbf{\begin{tabular}[c]{@{}c@{}}Camera\\ LiDAR\end{tabular}} & \textbf{S} & \textbf{C} & \textbf{S/C} & \textbf{AR} & \textbf{SS} & \textbf{VC} & \textbf{Mean} \\ \hline
\textbf{S}    & 83.86 & 86.74 & 88.25 & 87.64 & 87.70 & 88.54 & \cellcolor{green!25}85.61          \\
\textbf{C}    & 86.31 & 87.75 & 90.08 & 90.14 & 90.49 & 89.93 & \cellcolor{green!25}87.67          \\
\textbf{S/C}  & 88.05 & 90.32 & 89.81 & 92.92 & 90.16 & 92.08 & \cellcolor{green!25}89.38          \\
\textbf{AR}   & 87.27 & 89.59 & 91.36 & 92.94 & 92.55 & 91.52 & \cellcolor{green!25}88.99          \\
\textbf{SS}   & 85.52 & 90.01 & 88.41 & 92.55 & 94.79 & 90.24 & \cellcolor{green!25}87.86          \\
\textbf{VC}   & 87.08 & 89.19 & 90.95 & 90.94 & 92.33 & -     & \cellcolor{green!25}88.58          \\ \hline
\textbf{Mean} & 85.29 & 87.81 & 89.14 & 89.42 & 89.36 & 89.56 & \textbf{86.91}
\end{tabular}
\end{table}

The camera-LiDAR-based loop closure performs the best in all of the analyzed weather conditions when compared to camera-based and LiDAR-based solutions. 
Compared to camera-based loop closure, the presented version achieves the best gains in sunset conditions ($5.47$ percentage point increase in recognition rate) and sunny conditions ($3.56$ percentage point increase in recognition rate). 
This proves that additional LiDAR intensity data make loop closure more invariant to direct sunlight.

\begin{figure}[h!]
    \centering
    \includegraphics[width=\columnwidth]{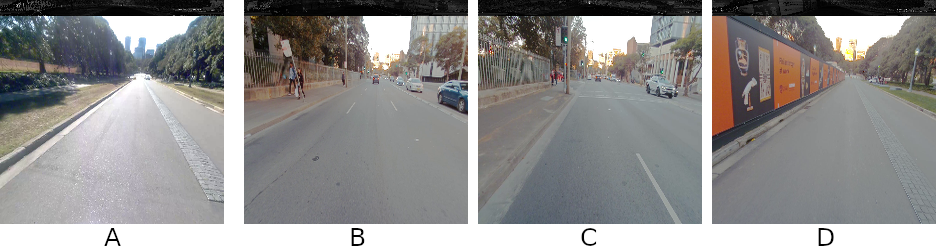}
    \caption{Examples of incorrect recognition for camera-LiDAR-based loop closure. 
    The input image (A) is incorrectly matched to two locations (B, C) based on the descriptor. The correct match (D) has the third-best descriptor match. The visual comparison proves that loop closure using a single image/scan is hard even for a person.}
    \label{fig:fusion2}
\end{figure}

We also took a closer look at cases when camera-LiDAR-based loop closure failed. An example of such a case is presented in Fig.~\ref{fig:fusion2} when a correct match had the third-best match to the input image based on the trained descriptor.
In this case, the loop closure was not recognized due to structural changes to the environment as the wall was no longer present in the input image.
Such real-world situations are hard to predict, but considering more than a single location descriptor could lead to an increase in the loop closure recognition rate.

\section{Conclusions}

We present a comparison of camera-based, LiDAR-based, and joint camera-LiDAR-based loop closures across varying weather conditions on the same trajectories using publicly available USyD dataset.
As the processing pipeline architectures for all considered solutions are the same, it is possible to conclude that the camera-based solution performance degrades in direct sunlight situations while LiDAR-based solution utilizing intensities provides overall similar performance independent of lighting conditions with worse performance in after the rain conditions.
These observations lead to the creation of a camera-LiDAR-based solution that performs best in all considered cases.

The presented experimental evaluation proves that multi-sensory loop closures should be considered in real-world scenarios as it provides more robust solution.
In our future work, we plan on utilizing pipelines more suited for the input from each sensor that could lead to the development of a system that can be directly compared with existing state-of-the-art camera-based and LiDAR-based loop closures.



\end{document}